\documentclass{article}

\usepackage{PRIMEarxiv}

\usepackage[utf8]{inputenc} 
\usepackage[T1]{fontenc}    
\usepackage{hyperref}       
\usepackage{url}            
\usepackage{booktabs}       
\usepackage{amsfonts}       
\usepackage{nicefrac}       
\usepackage{microtype}      
\usepackage{lipsum}
\usepackage{fancyhdr}       
\usepackage{graphicx}       
\graphicspath{{media/}}     

\usepackage{amsmath,amssymb}
\usepackage[ruled,vlined]{algorithm2e}
\usepackage{algorithm2e,setspace}
\usepackage{subfig}
\usepackage{multirow}
\usepackage{color}
\usepackage{cite}

\title{CLIP feature-based randomized control using images and text for multiple tasks and robots
\thanks{This is a preprint of an article submitted for consideration in ADVANCED ROBOTICS, copyright Taylor \& Francis and Robotics Society of Japan; ADVANCED ROBOTICS is available online at http://www.tandfonline.com/} 
}

\author{
  Kazuki Shibata, Hideki Deguchi and Shun Taguchi \\
  Collaborative Intelligence Research-Domain \\
  Toyota Central R\&D Labs., Inc. \\
  41-1, Yokomichi, Nagakute, Aichi, Japan \\
  \texttt{s-taguchi@mosk.tytlabs.co.jp} \\
}

\begin{document}
\maketitle

\begin{abstract}
This study presents a control framework leveraging vision language models (VLMs) for multiple tasks and robots. Notably, existing control methods using VLMs have achieved high performance in various tasks and robots in the training environment. However, these methods incur high costs for learning control policies for tasks and robots other than those in the training environment. Considering the application of industrial and household robots, learning in novel environments where robots are introduced is challenging. To address this issue, we propose a control framework that does not require learning control policies. Our framework combines the vision-language CLIP model with a randomized control. CLIP computes the similarity between images and texts by embedding them in the feature space. This study employs CLIP to compute the similarity between camera images and text representing the target state. In our method, the robot is controlled by a randomized controller that simultaneously explores and increases the similarity gradients. Moreover, we fine-tune the CLIP to improve the performance of the proposed method. Consequently, we confirm the effectiveness of our approach through a multitask simulation and a real robot experiment using a two-wheeled robot and robot arm.
\end{abstract}

\keywords{Vision-language model \and CLIP \and randomized control}

\section{Introduction}
In recent years, control methods using vision-language models (VLMs) have attracted attention in robotics and have been applied to navigation \cite{majumdar2022zson, dorbala2022clip, shah2022lmnav, gadre2022cows, zhou2023esc, huang2023visual, li2022envedit, huo2023geovln, li2023panogen} and manipulation \cite{mees2022calvin, xiao2023robotic, Chen2023, khandelwal2022simple, goodwin2022semantically, kapelyukh2023dall} tasks.
Because VLMs are trained using numerous images and texts on the web, they are expected to improve the generalization of control policies for various tasks using arbitrary texts as inputs that specify a task.



Existing control methods using VLMs have achieved high performance for various tasks and robots \cite{brohan2023rt1, brohan2023rt2, vuong2023open, li2023visionlanguage} in training environments. However, these methods incur high costs in learning control policies for tasks and robots that differ from the training environment. Consider a scenario in which a user utilizes a shipped robot at home or in a factory. In this scenario, the user cannot easily train the control policies in these environments after the robot is introduced. Therefore, it is crucial to construct a control framework that can reduce the cost of learning control policies.

\begin{figure}[!tp]
\begin{center}
\includegraphics[width=11cm]{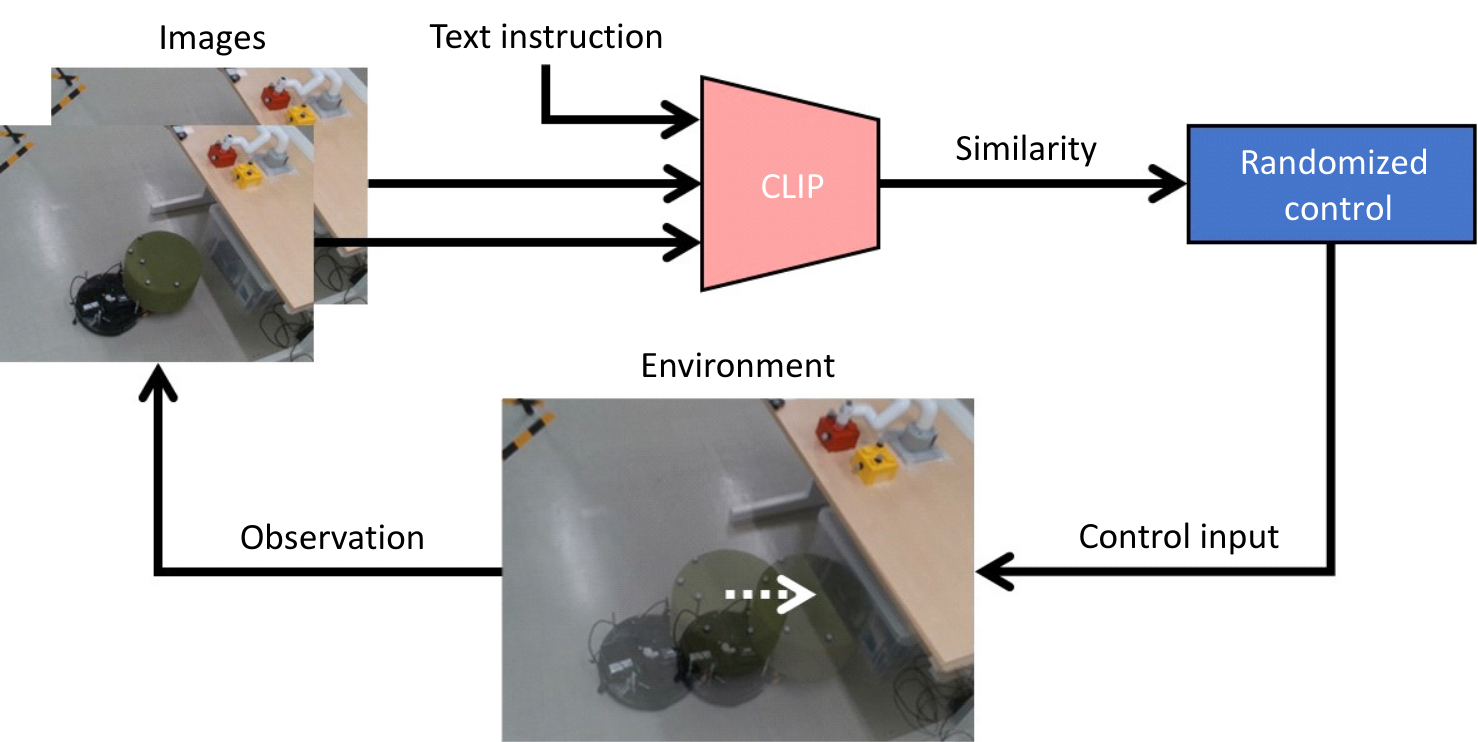}
\caption{Chair rearrangement task using CLIP feature-based randomized control. The text instruction is “place a green chair under the table.”}
\label{pull-fig}
\end{center}
\end{figure}

In this study, we proposed a control framework without learning control policies. Our framework combined the vision language model CLIP \cite{pmlr-v139-radford21a} with randomized control \cite{AZUMA20132307}, as shown in Figure \ref{pull-fig}. CLIP is a model trained using numerous images and text on the web. It computes the similarity between images and texts by embedding them into the feature space. In our method, the similarity between camera images and text representing the target state was computed using CLIP. The robot was controlled using a randomized control system that alternately repeated stochastic and deterministic movements. The former was used to compute the gradient of similarity, and the latter increased the similarity using the gradient. Moreover, we fine-tuned the CLIP to improve the performance of the proposed method. 
Although the proposed method commonly employs VLMs to be applicable for multiple tasks and robots, it differs from those in the literature \cite{brohan2023rt1, brohan2023rt2, vuong2023open, li2023visionlanguage} in that it does not require learning control policies.

To confirm the effectiveness of the proposed method, we conducted a multitask simulation using a robot arm and real experiment utilizing a two-wheeled robot and robot arm. We confirmed that the proposed method could be applied to multiple tasks in which a robot arm closes and opens a drawer, door, or window. Furthermore, we confirmed that the proposed method can be applied to different robots through a task wherein a two-wheeled robot places a chair under a table, and that in which a robot arm places a box next to another box.

The contributions of this study can be summarized as follows:
\begin{itemize}
\item We proposed a CLIP feature-based randomized control that can apply to multiple tasks and robots without learning control policies.
\item We confirmed the generalization of the proposed method for multiple tasks via a multitask simulation using a robot arm.
\item We verified the generalization of the proposed method for different robots via a real experiment using a two-wheeled robot and robot arm.
\end{itemize}

The remainder of this paper is organized as follows:
Section 2 introduces related work.
Section 3 outlines robot control strategies using images and text for multiple tasks.
Section 4 presents multitask simulation results utilizing a robot arm.
Section 5 describes a real experiment using a two-wheeled robot and robot arm.
Section 6 discusses the limitations of this study and the scope of future work.
Finally, Section 7 summarizes the study.

\section{Related work}
In recent years, VLMs have attracted attention in the field of robotics and have been applied to various tasks, including object-goal navigation \cite{majumdar2022zson, dorbala2022clip, shah2022lmnav, gadre2022cows, zhou2023esc, huang2023visual}, room-to-room navigation \cite{li2022envedit, huo2023geovln, li2023panogen}, pick and place \cite{mees2022calvin, xiao2023robotic, Chen2023}, and rearrangement tasks \cite{khandelwal2022simple, goodwin2022semantically, kapelyukh2023dall}.
Leveraging the numerous images and texts on the web, VLMs are expected to enhance the generalization of control policies for extensive tasks.

Several studies employing VLMs have succeeded in generalizing multiple tasks and robots \cite{brohan2023rt1, brohan2023rt2, vuong2023open, li2023visionlanguage}. For example, Brohan et al. \cite{brohan2023rt1} proposed Robotics Transformer 1 (RT-1), which is a vision-language-action model based on transformer architecture. RT-1 inputs camera images and task instruction text, encodes them as tokens using a pre-trained FiLM EfficientNet model \cite{tan2019efficientnet}, and compresses them using TokenLearner \cite{ryoo2021tokenlearner}. The compressed tokens are input to the transformer, which outputs the action tokens. Training on extensive real manipulation data enables applications in various tasks in real-world environments. Moreover, based on RT-1, they introduced Robotics Transformer 2 (RT-2) \cite{brohan2023rt2}, which employs VLMs such as PaLI-X \cite{chen2023pali} and PaLM-E \cite{driess2023palme}. Training on both real manipulation and web data improves generalization performance for unknown tasks better than RT-1. Furthermore, Vuong et al. \cite{vuong2023open} proposed the Robotic Transformer X (RT-X), which can be applied to different types of robots. They presented two model architectures, RT-1-X and RT-2-X, which employed RT-1 and RT-2, respectively, and trained these models using data collected from 22 types of robot arms. Although these methods exhibit high performance for extensive tasks and robots in training environments, they incur high costs for learning control policies for tasks and robots other than the training environments.

The proposed framework combines a vision-language model, CLIP, with randomized control \cite{AZUMA20132307} for multiple tasks and robots without learning control policies. To the best of our knowledge, no similar work using randomized controls with VLMs has been reported so far. Although the present study has commonly used VLMs that can be applied to multiple robots and tasks, it differs from \cite{brohan2023rt1, brohan2023rt2, vuong2023open, li2023visionlanguage} in that our method does not require learning control policies. Moreover, we confirmed the generalization of our method for different types of robots, including a two-wheeled robot and robot arm, whereas other studies were limited to robot arms.

\section{Robot control using images and text for multiple tasks and robots}
\subsection{Problem setting}
This section describes the problem setting for robot control using images and text.
The environment was observed using a camera at a fixed position and orientation. Moreover, text instructions were used to enable the robot to perform tasks. For example, in a chair-rearrangement task, the text instruction is given as “place a green chair under the table.”
In this study, we assumed that the position and orientation of the robot and position of the object can be observed, and these values were assigned to the robot as inputs. 

This study aims to control the object position to reach the target position for multiple tasks and robots.

\subsection{Overview of our control framework}
\begin{figure}[!tp]
\begin{center}
\includegraphics[width=14cm]{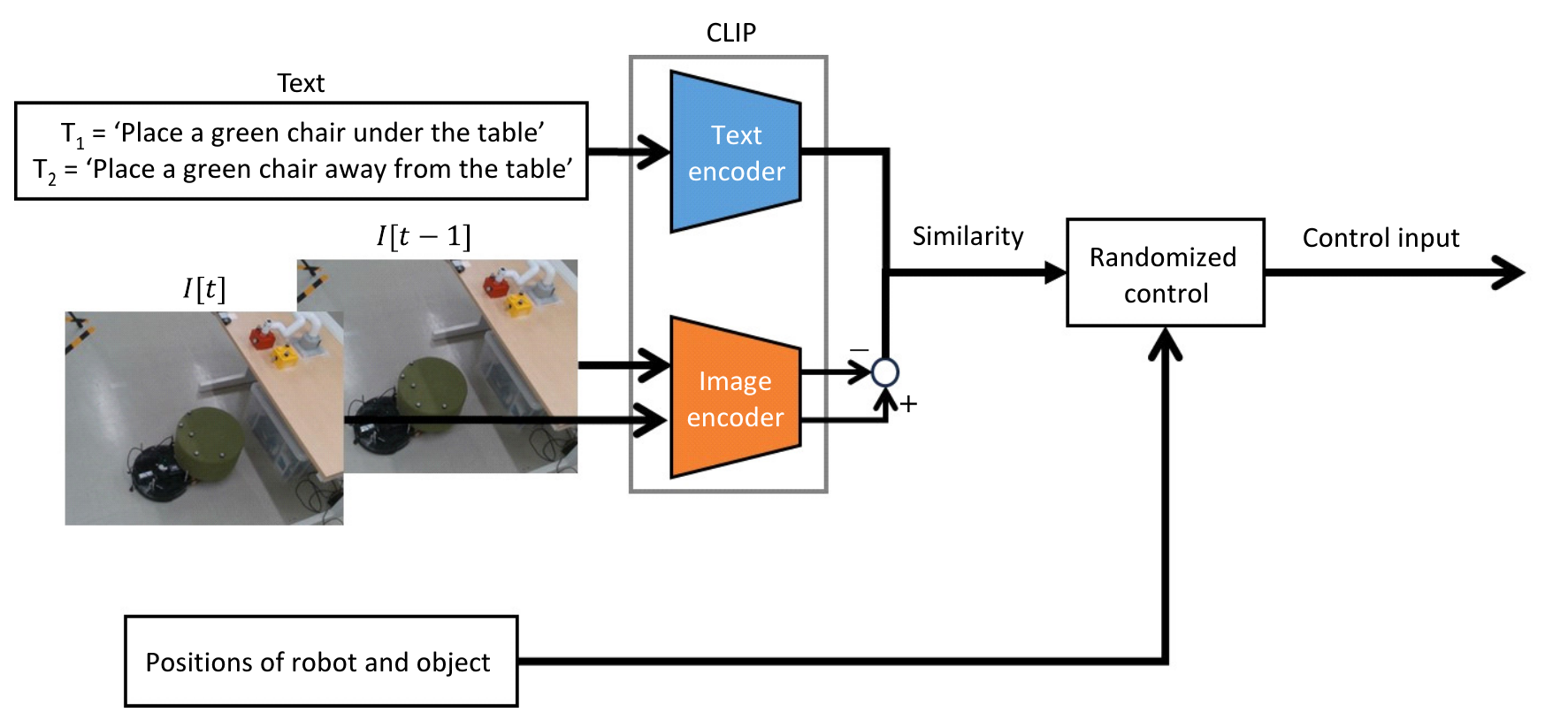}
\caption{Overview of our control framework}
\label{framework}
\end{center}
\end{figure}

An overview of the control framework is presented in Figure \ref{framework}.
In our framework, the robot computes a similarity that determines whether the difference between two image features is close to the instruction text or text representing an action opposite to the instruction using CLIP.
We used the difference between two image features because object motions such as open and close cannot easily be determined using a single image.


The robot inputs the positions of the robot and object, image $I[t]$ at the current control step $t$, image $I[t-1]$ at the control step $t-1$, instruction text $T_1$, and text representing the opposite action to instruction $T_2$. The robot computes the image features $\textit{\textbf{h}}[t]$ and $\textit{\textbf{h}}[t-1]$ using the CLIP image encoder.
Moreover, the robot computes the text features $\textit{\textbf{g}}_1$ and $\textit{\textbf{g}}_2$ of $T_1$ and $T_2$ using the CLIP text encoder.
The robot computes the similarity $R_i[t]$ ($i=1,2$) between $\textit{\textbf{h}}[t]-\textit{\textbf{h} }[t-1]$ and $\textit{\textbf{g}}_i$. The control input is computed such that the similarity difference $R_1[t]-R_2[t]$ is positive.

\subsection{CLIP feature-based randomized control}
This subsection describes the details of our CLIP feature-based randomized control.
To generate motion from two images, we compute the cosine similarity of $\textit{\textbf{h}}[t]-\textit{\textbf{h}}[t-1]$ and $\textit{\textbf{g}}_i$ ($i=1,2$) using
\begin{align}
R_i[t]=\frac{(\textit{\textbf{h}}[t]-\textit{\textbf{h}}[t-1])^{\top}\cdot \textit{\textbf{g}}_i}{\|\textit{\textbf{h}}[t]-\textit{\textbf{h}}[t-1]\|\|\textit{\textbf{g}}_i\|}
\label{similarity}
\end{align}
Furthermore, the similarity gradient is computed using
\begin{align}
dV_i[t]=\frac{R_1[t]-R_2[t]}{x_{i,\rm R}[t]-x_{i,\rm R}[t-1]}
\label{grad}
\end{align}
where $x_{i,\rm R}[t]$ represents the position of the robot along the axis $i$ at control step $t$. 

Next, we introduce our CLIP feature-based randomized control method, which increases the similarity between images and text.
Randomized control \cite{AZUMA20132307} is a control law that maximizes an unknown evaluation function by alternately repeating stochastic and deterministic movements. The former computes the gradient of the evaluation function, and the latter increases the evaluation function using the gradient.
Using this control law, the control input is computed as follows:
\begin{align}
u_i[t]=
\begin{cases}
c\Delta_i[t] \ &{\rm if}\ t\ {\rm is\ odd} \\
f(dV_i[t])\ &{\rm otherwise}
\end{cases},
\label{ctrl}
\end{align}
where $i$ is the $x$, $y$, or $z$ axis; $\Delta_i[t]$ is a random variable that randomly becomes $+1$ or $-1$ with a probability of 0.5, $c>0$; and $f$ is a function that determines the updated amount using the gradient.
If $t$ is an odd number, the robot selects a stochastic movement in each axis direction to compute the similarity gradient. If $t$ is an even number, the robot selects a deterministic movement by using a gradient to increase the similarity.
In \cite{AZUMA20132307}, the control input is calculated using $f(dV_i[t])=kdV_i[t]$ ($k>0$).
However, if $dV_i[t]$ changes significantly, the position of the robot may vibrate.

In this study, the function $f$ is computed using RMSprop \cite{hinton2012neural} as follows:
\begin{align}
f(dV_i[t])&=\alpha \frac{dV_i[t]}{\sqrt{\hat{v}[t]}+\epsilon}, \label{rmsprop1} \\
v_i[t]&=\beta v_i[t-1]+(1-\beta)dV_i[t]^2, \label{rmsprop2}
\end{align}
where $\alpha>0$, $\epsilon>0$, and $0\le \beta<1$.
The moving average in (\ref{rmsprop2}) was employed to prevent the vibration of the robot, even if $dV_i[t]$ changed significantly. We adopted RMSprop because RMSprop achieved a better performance than Adam \cite{kingma2014adam} when applying randomized control.

As our method calculates control inputs using the similarity of CLIP features by alternately repeating stochastic and deterministic movements, it can be applied to different robots without learning control policy.

\subsection{Fine-tuning of the CLIP model for multiple tasks}
This section introduces the fine-tuning of the CLIP model, which can be applied to multiple tasks. Although the original CLIP model is suitable for classifying the names of objects, it is not appropriate for classifying object motions such as open and close. Therefore, we aim to fine-tune the CLIP model to classify the object motions for multiple tasks. The fine-tuning algorithm is presented in Algorithm \ref{algo}.

The first step is the collection of data for multiple tasks. 
For task $i$ ($i=1,\cdots,N$), data $\mathcal{D}^i[k]=\{I[k],y[k]\}$ at control step $k$, where $y[k]$ is a negative value of the distance between the object position and its target position, as follows:
\begin{align}
y[k]=-\|\textit{\textbf{r}}_o-\textit{\textbf{x}}_o[k]\|,
\end{align}
where $\textit{\textbf{r}} _o\in \mathbb{R}^3$ represents the target position of the object, and $\textit{\textbf{x}}_o\in \mathbb{R}^3$ indicates its current position.
In addition, to collect images when the object motion changes, data are collected if $\|y[k]-y[k-1]\|>\delta_y$ is satisfied, where $\delta_y$ is the threshold for data collection.
Data collection was repeated until the number of data reached a certain value $M$ for each task.

From the collected data $\mathcal{D}^i=\{\mathcal{D}^i[1],\cdots,\mathcal{D}^i[M]\}$, $\mathcal{D}_1^i=\{I_1, y_1\}$, and $\mathcal{D}_2^i=\{I_2,y_2\}$ are randomly sampled.
The ground truth of the text is given by
\begin{align}
T^{\ast}=[T^{\ast}_1,T^{\ast}_2]=
\begin{cases}
[T_1,T_2] \ &{\rm if}\ y_2>y_1 \\
[T_2,T_1] &{\rm otherwise}
\end{cases}.
\label{text}
\end{align}

Using the CLIP features $\textit{\textbf{h}}_1$ and $\textit{\textbf{h}}_2$ of images $I_1$ and $I_2$, the difference in CLIP features $\textit{ \textbf{h}}^-=[\textit{\textbf{h}}_2-\textit{\textbf{h}}_1, \textit{\textbf{h}}_1-\textit{\textbf{h}}_2]$ is computed.
For text $T^{\ast}$, the CLIP feature $\textit{\textbf{g}}^{\ast}=[\textit{\textbf{g}}_1^{\ast} ,\textit{\textbf{g}}_2^{\ast}]$ is computed.

Finally, the cosine similarity is calculated for each element of $\textit{\textbf{h}}^-$ and $\textit{\textbf{g}}^{\ast}$, and the CLIP model is updated to minimize the cross-entropy loss.
See \cite{pmlr-v139-radford21a} for details of the optimization.

\begin{algorithm}[!tp]
\# Data collection for $N$ tasks \\
\For{$ i = 1, ..., N$} 
{
	Set $t$ = 1 \\
	\For{$ k = 1, ...$}
	{
		Collect $\mathcal{D}^i[t]=\{I[k],y[k]\}$ if $\|y[k]-y[k-1]\|>\delta_y$\\
		$t\leftarrow t+1$ \\
		if $t=M$ break
	}
}
\# Fine-tuning of CLIP model \\
\For{$ i = 1, ..., N$}
{
    Set $T_1$ and $T_2$ \\
    \For{$ k = 1, ..., K$}
    {
	Sample $\mathcal{D}_1^i$ and $\mathcal{D}_2^i$ from $\mathcal{D}^i=\{\mathcal{D}^i[1],\cdots,\mathcal{D}^i[M]\}$ \\
	Compute ground-truth text $T^{\ast}$ using (\ref{text}) \\
	Compute image features $\textit{\textbf{h}}_1$ and $\textit{\textbf{h}}_2$ using an image encoder \\
        Compute $\textit{\textbf{h}}^-=[\textit{\textbf{h}}_2-\textit{\textbf{h}}_1, \textit{\textbf{h}}_1-\textit{\textbf{h}}_2]$ \\
        Compute $\textit{\textbf{g}}^{\ast}$ using $T^{\ast}$ with a text encoder \\
        Compute loss using $\textit{\textbf{h}}^-$ and $\textit{\textbf{g}}^*$ and update the CLIP model according to \cite{pmlr-v139-radford21a} \\
    }
}
\caption{Fine-tuning of the CLIP model for multiple tasks}
\label{algo}
\end{algorithm}

\section{Simulation}
In this section, we confirm the generalization of the proposed method for multiple tasks through a multitask simulation using a robot arm. This simulation is aimed to confirm that the presented method can perform multiple tasks even if the target position of the object is unknown.

\subsection{Simulation environment}
\begin{figure}
\centering
\subfloat[drawer-close]{
\includegraphics[height=3.75cm,width=4cm]{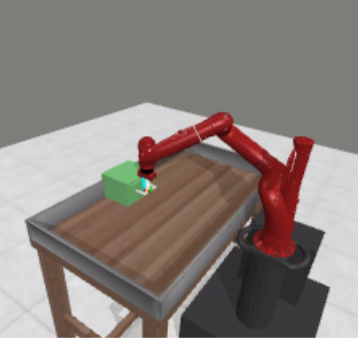}
\label{drawer-close}}
\subfloat[drawer-open]{
\includegraphics[height=3.75cm, width=4cm]{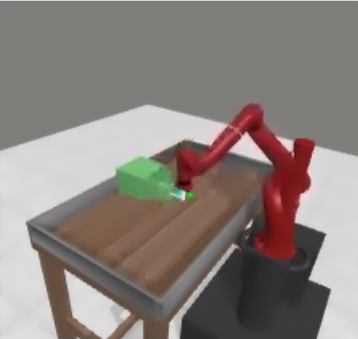}
\label{drawer-open}}
\subfloat[door-close]{
\includegraphics[height=3.75cm, width=4cm]{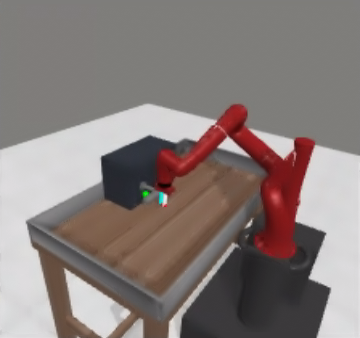}
\label{door-close}}
\\
\subfloat[door-open]{
\includegraphics[height=3.75cm, width=4cm]{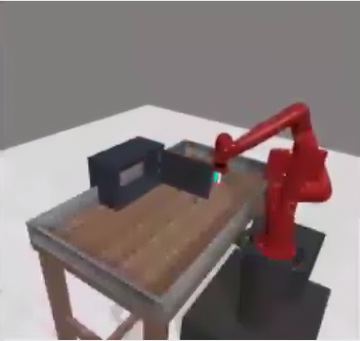}
\label{door-open}}
\subfloat[window-close]{
\includegraphics[height=3.75cm, width=4cm]{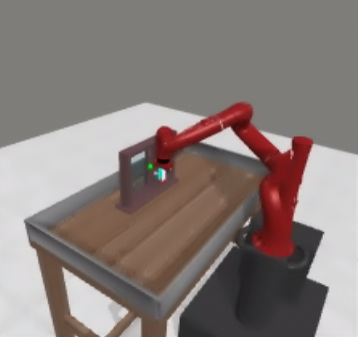}
\label{window-close}}
\subfloat[window-open]{
\includegraphics[height=3.75cm, width=4cm]{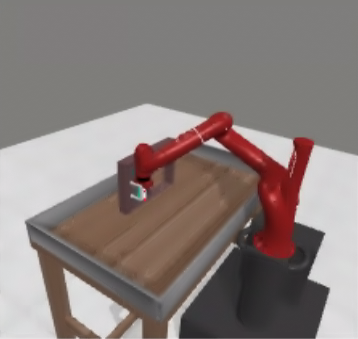}
\label{window-open}}
\caption{Simulation environment (the green dot indicates the target position of the handle)}
\label{task}
\end{figure}

The simulation was performed using a simulator Metaworld \cite{pmlr-v100-yu20a}, which can verify various tasks using a robot arm. The simulation environment is shown in Figure \ref{task}. 
The success of each task is determined as follows:
\begin{itemize}
\item drawer-close/open: the distance between the position of the drawer handle and its target position is within 0.03 m.
\item door-close/open: the distance between the position of the door handle and its target position is within 0.05 m.
\item window-close/open: the distance between the position of the window handle and its target position is within 0.05 m.
\end{itemize}

\subsection{Control implementation}
The control time step was set to 0.1 s, and the total number of steps was set to 5.0$\times 10^2$ for window-close and window-open tasks and 1.0$\times 10^3$ for other tasks. Moreover, we set $c=0.2$, $\alpha=1.0$, $\beta=0.5$, and $\epsilon=1.0\times 10^{-8}$.
These parameters were set such that the robot arm could reach the target position of the end effector while avoiding overshooting.

In the simulation, the positions of the robot and object were considered as those of the end effector and handle, respectively.
To make the end effector approach the handle, we add the distance between the end effector and handle to (\ref{grad}) as follows:
\begin{align}
dV_i[t]\leftarrow dV_i[t]-\frac{\|x_{i,\rm o}[t]-x_{i,\rm R}[t]\|^2-\|x_{i,\rm o}[t-1]-x_{i,\rm R}[t-1]\|^2}{x_{i,\rm R}[t]-x_{i,\rm R}[t-1]}, 
\label{grad2}
\end{align}
where $x_{i,\rm o}[t]\in \mathbb{R}$ represents the handle position along the $i$-axis.

Furthermore, to prevent the gradient from changing significantly, we adopted the upper and lower limits in (\ref{grad}) as follows:
\begin{align}
dV_i(t) \leftarrow
\begin{cases}
-\lambda_i &{\rm if}\ dV_i(t) <-\lambda_i \\
+\lambda_i &{\rm if}\ dV_i(t) >+\lambda_i \\
dV_i(t) &{\rm otherwise} \\
\end{cases},
\label{grad3}
\end{align}
where we set $[\lambda_x,\lambda_y,\lambda_z]=[1.0,1.0,0.1]$ in the simulations.

Moreover, the robot sometimes encountered a stuck where the handle did not move, even if it moved the end effector when it was sufficiently close to the handle. To address this issue, we replace the position of the object with 
\begin{align}
\textit{\textbf{x}}_{\rm o}[t]\leftarrow
\begin{cases}
\textit{\textbf{x}}_o[t] + [0, 0, 0.05]^{\top} \ &{\rm if}\ d_e[t]<\delta_e \ \& \ \|\textit{\textbf{x}}_o[t]-\textit{\textbf{x}}_o[0]\|<\delta_o \\
\textit{\textbf{x}}_o[t] \ &{\rm otherwise}
\end{cases},
\label{des-pos}
\end{align}
where the first condition determines that the end effector is stuck if the travel distance of the end effector $d_e[t]$ in the last 1.0 s is less than a certain threshold $\delta_e$.
The second condition determines whether the end effector is stuck if the distance between the initial and current positions of the handle is less than a certain threshold $\delta_o$.
If the two conditions are satisfied, the robot arm avoids being stuck by setting the target position to be a point, which is located 0.05 m above the handle position.
In the simulation, we set $\delta_e=0.02$ m and $\delta_o=0.05$ m.

To confirm the effectiveness of the proposed method, we compared it with our method with a reinforcement-learning algorithm PPO \cite{schulman2017proximal}, which is a deep actor-critic algorithm. The inputs of the policy are set as $\left[i/N,\textit{\textbf{x}}_o[t], \textit{\textbf{x}}_{\rm R}[t], \textit{\textbf{x}}_o[t-1], \textit{\textbf{x}}_{\rm R}[t-1]\right]$, where $i$ ($i=1,\cdots,N$) is a task ID, and the target position of the handle is unknown. The reward function is used as described in \cite{pmlr-v100-yu20a}. We trained the same policy for six tasks by using the code in \cite{pytorch_minimal_ppo}. The critic and actor networks contained two hidden layers of 128 units, respectively. The activation function of the output layer in the critic network is linear, whereas that in the actor network is $\tanh$. The hyperparameters used in the PPO algorithm are presented in Table \ref{PPOhyperparameters}.

Moreover, we evaluated several control methods using the proposed method. To evaluate performance under ideal conditions, where the robot knows the target position of the handle, we evaluated a control method by setting the first term in (\ref{grad2}) to 
\begin{align}
-\frac{\|r_{i,o}-x_{i,o}[t]\|^2-\|r_{i,o}-x_{i,o}[t-1]\|^2}{x_{i,o}[t]-x_{i,o}[t-1]},
\end{align}
where $r_{i,o}$ represents the target position of the handle on the $i$-axis. This control method is denoted by the \textbf{Goal}. Moreover, we evaluated several control methods using the CLIP models ViT-B/32 and ViT-L/14, which employ a Vision Transformer \cite{dosovitskiy2021image} with a higher performance than ResNet \cite{he2015deep}.
In addition, we evaluated a control method using fine-tuned ViT-B/32, which was trained on the same model for multiple tasks, using the method described in Subsection 3.3.
These methods are denoted as \textbf{ViT-B/32}, \textbf{ViT-L/14}, and \textbf{ViT-B/32 (finetune)}.

The texts used in the simulation are presented in Table \ref{prompt}. These texts were determined through prompt engineering to improve the performances of ViT-B/32 and ViT-L/14.

\begin{table}
\small
\caption{Hyperparameters used in the PPO algorithm}
\vspace{1mm}
  \label{PPOhyperparameters}
  \centering
  \renewcommand{\arraystretch}{1.2}
  \begin{tabular}{cc}
    \hline
     Parameter & Value \\
    \hline \hline
     Number of episodes & 3.0$\times 10^3$ \\ 
     Discount factor & 0.99 \\ 
     Clipping parameter & 0.2 \\
     Number of episodes for update & 80 \\
     Learning rate for the critic network & 3.0$\times 10^{-4}$  \\ 
     Learning rate for the actor network  & 1.0$\times 10^{-3}$  \\      
    \hline     
  \end{tabular}
\end{table}

\begin{table}[!tp]
\centering
\caption{Texts used in a multitask simulation}
\vspace{1mm}
\label{prompt}
\centering
\small
\begin{tabular}{ccc}
\hline
Task & Variable & Text \\
\hline \hline
\multirow{2}{*}{drawer-close} & $T_1$ & close a drawer with a drawer handle \\ 
& $T_2$ & open a drawer with a drawer handle \\ 
\hline
\multirow{2}{*}{drawer-open} & $T_1$ & open a drawer with a drawer handle \\ 
& $T_2$ & close a drawer with a drawer handle \\ 
\hline
\multirow{2}{*}{door-close} & $T_1$ & close a door with a door handle \\ 
& $T_2$ & open a door with a door handle \\ 
\hline     
\multirow{2}{*}{door-open} & $T_1$ & open a door with a door handle \\ 
& $T_2$ & close a door with a door handle \\ 
\hline
\multirow{2}{*}{window-close} & $T_1$ & close a window in the right direction \\ 
& $T_2$ & open a window in the left direction \\ 
\hline
\multirow{2}{*}{window-open} & $T_1$ & open a window in the left direction \\ 
& $T_2$ & close a window in the right direction \\ 
\hline  
\end{tabular}
\end{table}

\subsection{Result}
Table \ref{sim-eval1} compares the success rates of applying each method to six tasks for 100 trials.
Notably, the target positions of the handles are unknown except for \textbf{Goal}.
The results showed that the PPO method could not perform the drawer-open and door-open tasks. In these tasks, the PPO method could make the end effector approach the handle but could not move the handle toward the goal.
Meanwhile, \textbf{ViT-B/32} and \textbf{ViT-L/14} achieved higher success rates than the PPO method.
Furthermore, \textbf{ViT-B/32 (finetune)} achieved higher success rates than \textbf{ViT-B/32} and \textbf{ViT-L/14} for various tasks, indicating that the performance of our method could be improved by fine-tuning the CLIP model.
Furthermore, the number of collected images for fine-tuning the CLIP is 1.0$\times10^5$ while those for training the policy using PPO is 1.5$\times10^6$. This result indicates that the proposed method requires fewer images than PPO.

To investigate the factors that \textbf{ViT-B/32 (finetune)} achieved a high success rate, we evaluated the accuracy rate of the text for the two images. This is because the higher the accuracy rate of the text, the more accurate the sign of the gradient in (\ref{grad}), which can increase the similarity.
The accuracy rate of the text is defined as
\begin{align}
A&=\frac{1}{N_t}\sum^{N_t}_{t=1} a_t, \label{accuracy1} \\
a_t&=
\begin{cases}
1 \ &{\rm if}\ (R_1-R_2)(y_1-y_2)\ge0 \\
0 \ &{\rm otherwise}
\end{cases},
\label{accuracy2}
\end{align}
where $N_t$ in (\ref{accuracy1}) is the total number of trials, and $R_i$ ($i=1,2$) in (\ref{accuracy2}) is computed using (\ref{similarity}) for the two randomly sampled images.

Table \ref{sim-eval2} compares the text accuracy rates for $2.0\times 10^3$ trials when each CLIP model is applied to the 2.0$\times 10^3$ images.
The results show that \textbf{ViT-B/32 (finetune)} achieved higher text accuracy rates than \textbf{ViT-B/32} and \textbf{ViT-L/14} for all tasks.
Furthermore, the text accuracy and success rate of \textbf{ViT-L/14} were higher than those of \textbf{ViT-B/32} for all the tasks, as presented in Tables \ref{sim-eval1} and \ref{sim-eval2}.
These results indicated that the higher the text accuracy rate, the higher the success rate.

Overall, we confirmed that the proposed method can control an object for multiple tasks without learning the control policy. Moreover, we improved the performance of our method by fine-tuning the CLIP model through a multitask simulation.

\begin{table}
\small
\centering
\caption{Comparison of success rates for the multitask simulation (the target position of the handle is unknown except for $\dagger$)}
\vspace{1mm}
  \label{sim-eval1}
  \renewcommand{\arraystretch}{1.2}
  \begin{tabular}{ccccccc}
    \hline
     Task & PPO & \textbf{Goal}$\dagger$ & \textbf{ViT-B/32} & \textbf{ViT-L/14} & \textbf{ViT-B/32 (finetune)} \\
    \hline \hline
     drawer-close & 0.80 & (0.93) & 0.59  & 0.66  & \textbf{0.93} \\ 
     drawer-open  & 0.00  & (0.98) & 0.22  & 0.49  & \textbf{0.94} \\ 
     door-close   & 0.96 & (0.99) & 0.96  & 0.98  & \textbf{0.99} \\ 
     door-open    & 0.00  & (0.89) & 0.17 & 0.21  & \textbf{0.82} \\ 
     window-close & \textbf{1.00} & (0.96) & 0.93 & 0.93 & 0.93 \\
     window-open  & \textbf{0.91} & (0.93)  & 0.87 & 0.89 & 0.90 \\ 
    \hline     
  \end{tabular}
\end{table}

\begin{table}
\small
\centering
\caption{Comparison of text accuracy rates for the simulator images}
\vspace{1mm}
  \label{sim-eval2}
  \centering
  \renewcommand{\arraystretch}{1.2}
  \begin{tabular}{cccc}
    \hline
     Task & \textbf{ViT-B/32} & \textbf{ViT-L/14} & \textbf{ViT-B/32 (finetune)} \\
    \hline \hline
     drawer-close & 0.50 & 0.73  & \textbf{0.94} \\ 
     drawer-open  & 0.49 & 0.69  & \textbf{0.95} \\ 
     door-close   & 0.55 & 0.73  & \textbf{0.97} \\ 
     door-open    & 0.63 & 0.78  & \textbf{0.94} \\ 
     window-close & 0.52 & 0.71  & \textbf{0.90} \\ 
     window-open  & 0.68 & 0.73  & \textbf{0.91} \\ 
    \hline     
  \end{tabular}
\end{table}

\section{Real robot experiment}
In this section, we confirm the generalization of the proposed method for different robots through real robot experiments using a two-wheeled robot and robot arm. This experiment aims to confirm that our method can control objects for different robots, even if the target position of the object is unknown.

\subsection{Experimental configuration}
\begin{figure}
\begin{center}
\includegraphics[width=9cm]{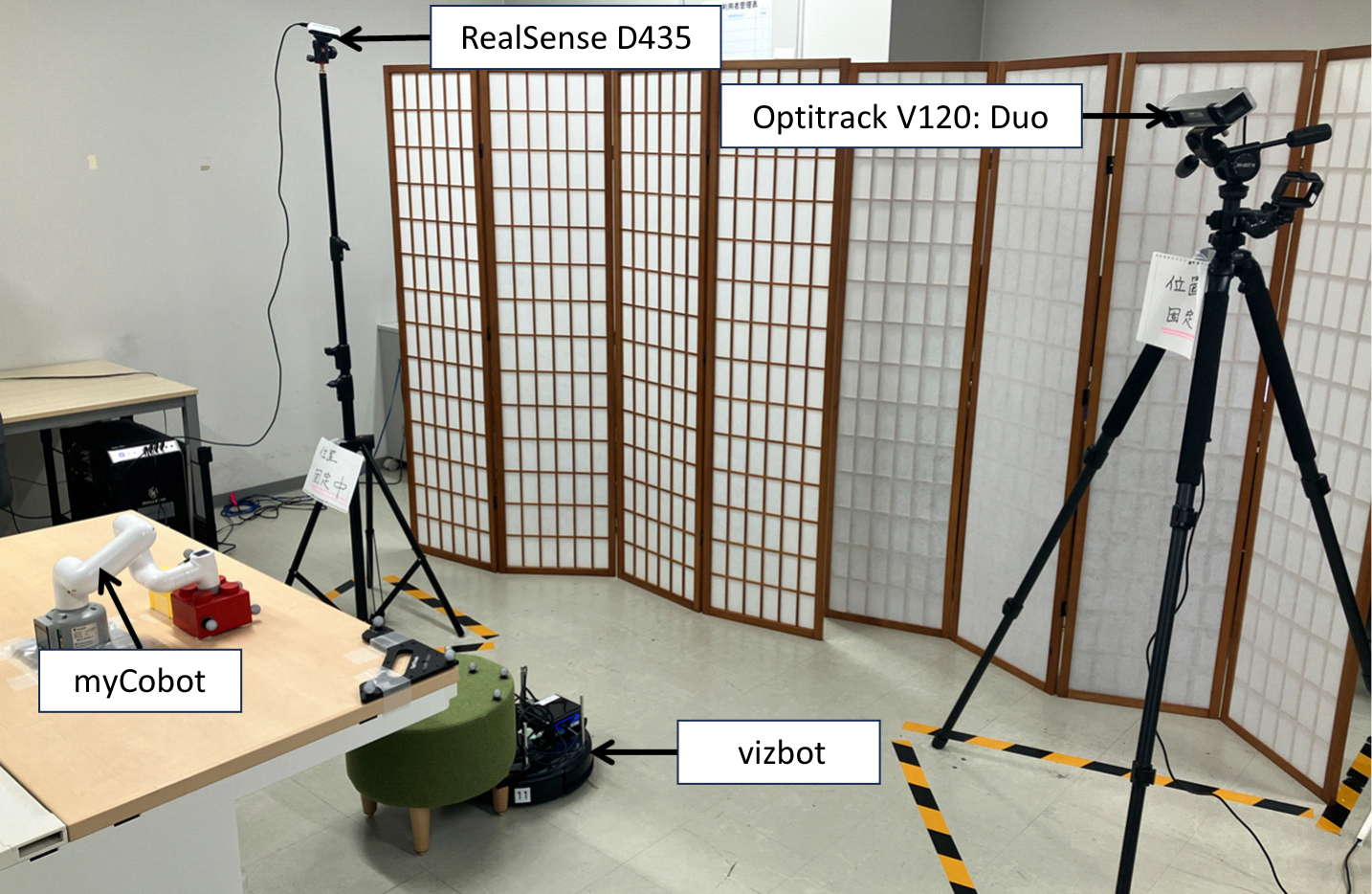}
\caption{Experimental configuration}
\label{config}
\end{center}
\end{figure}

The experimental configuration is shown in Figure \ref{config}.
The position of the object was observed using motion capture Optitrack V120: Duo at 100 fps. 
We set the position of the robot to be the same as that of the object because the robot was rigidly attached to the object.
Images were captured using an Intel RealSense D435 camera at 30 fps with a fixed position and angle.

To confirm the effectiveness of our method, we performed two rearrangement tasks using a two-wheeled robot vizbot \cite{Niwa2022} and a robotic arm, myCobot 280 M5 (Elephant Robotics Co., Ltd.).
The tasks are described as follows:
\begin{itemize}
\item chair-rearrangement: the vizbot places a green chair under a table. Success is determined when the position of the chair is inside the edge of the table.
\item box-rearrangement: the myCobot, which is fixed on the table, places a red box next to a yellow box. Success is determined when the position of the red box is within 0.05 m from the position exactly adjacent to the yellow box.
\end{itemize}

\subsection{Control implementation}
The control time step and total number of steps were set to 0.2 s and $2.0\times10^2$ steps, respectively, for the vizbot and these values were set to 1.0 s and $5.0\times10^1$ steps for the myCobot. 
The control inputs are the linear and angular velocity inputs for vizbot and the amount of movement of the end effector in the $x$- and $y$-axes for myCobot.
We set $c=0.1$ and $c=0.02$ for vizbot and myCobot while setting the parameters of RMSprop similar to the simulation. These parameters were set so that the robots could allow the object to reach its target position while avoiding overshooting.
The texts used in the experiment are shown in Table \ref{prompt2}.

\begin{table}[!tp]
\caption{Texts used in the experiment}
\small
\vspace{1mm}
\label{prompt2}
\centering
\begin{tabular}{ccc}
\hline
Task & Variable & Text \\
\hline \hline
\multirow{2}{*}{chair-rearrangement} & $T_1$ & place a green chair under the table \\ 
& $T_2$ & place a green chair away from the table \\ 
\hline
\multirow{2}{*}{box-rearrangement} & $T_1$ & place a red box next to the yellow box \\ 
& $T_2$ & place a red box away from the yellow box \\ 
\hline
\end{tabular}
\end{table}

We evaluated several control methods, \textbf{Goal}, \textbf{ViT-B/32}, \textbf{ViT-L/14}, and \textbf{ViT-B/32 (finetune)}, as described in the simulation.
In the experiment, we collected 4.0$\times 10^3$ data for each robot via manual control, and the CLIP model was trained on the same model for the robot arm and two-wheeled robot.

\subsection{Result}
Table \ref{eval1} compares the success rates when applying each method to 30 trials, where we set five different initial positions and performed six experiments for each.
The target positions of the objects are unknown except for \textbf{Goal}.
The results show that \textbf{ViT-B/32} and \textbf{ViT-L/14} could perform both tasks in several trials without fine-tuning the CLIP model. 
Furthermore, \textbf{ViT-B/32 (finetune)} achieved higher success rates than \textbf{ViT-B/32} and \textbf{ViT-L/14} for both tasks, indicating that the performance of our method could be improved by fine-tuning the CLIP model.

Table \ref{eval2} compares the text accuracy rates for $2.0\times 10^3$ trials when applying each CLIP model to $2.0\times 10^3$ images.
The results showed that \textbf{ViT-B/32 (finetune)} achieved higher accuracy rates than the other methods for both tasks. Furthermore, \textbf{ViT-L/14} achieved a higher text accuracy and success rate than \textbf{ViT-B/32} for the box-arrangement task, whereas \textbf{ViT-B/32} achieved these values higher than \textbf{ViT-L/14} for the chair-rearrangement task.
These results indicate that the higher the text accuracy rate, the higher the success rate.

\begin{figure}[!tp]
\centering
\subfloat[\textbf{ViT-B/32}]{
\includegraphics[width=15cm]{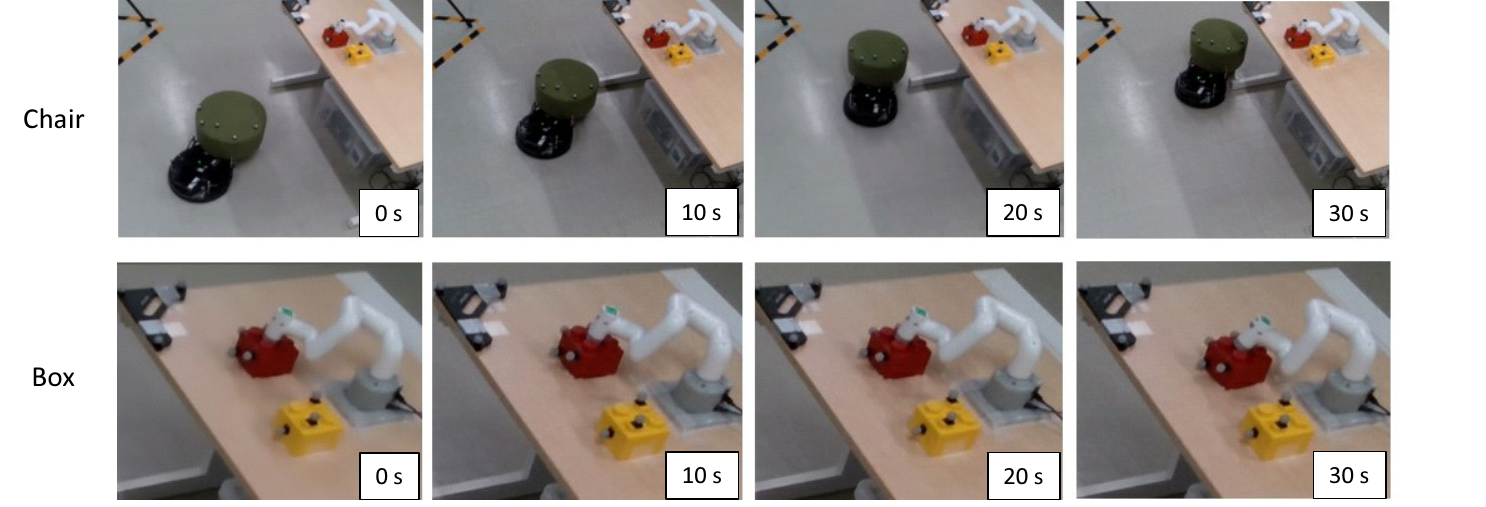}
\label{result-ViTB32}}
\\
\subfloat[\textbf{ViT-L/14}]{
\includegraphics[width=15cm]{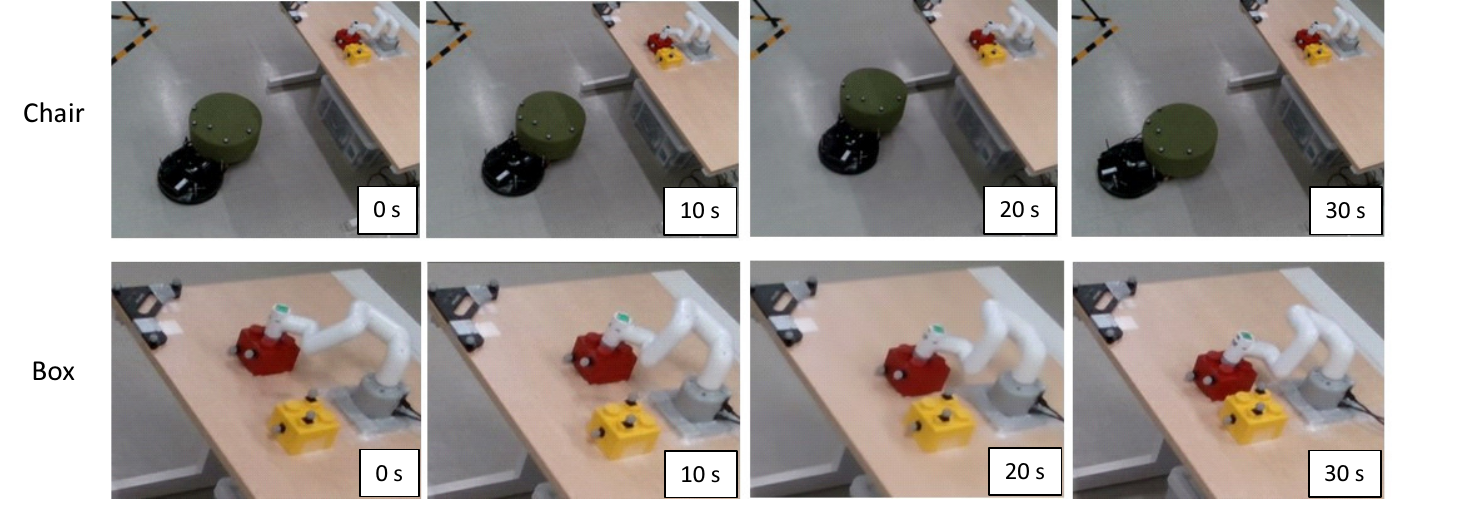}
\label{result-ViTL14}}
\\
\subfloat[\textbf{ViT-B/32 (finetune)}]{
\includegraphics[width=15cm]{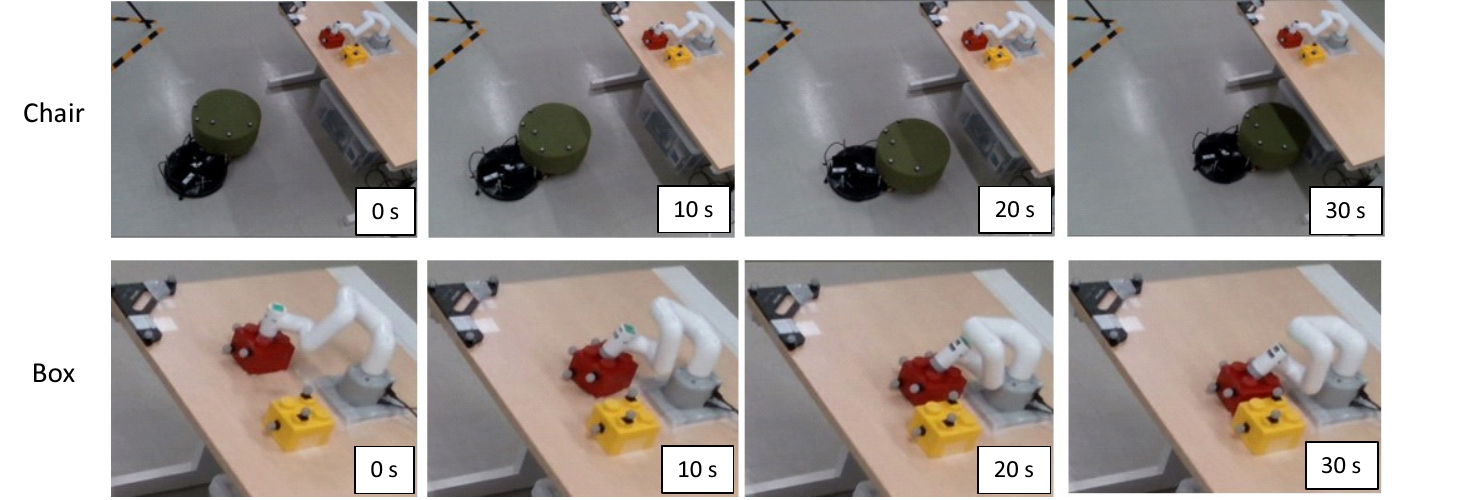}
\label{result-ours}}
\caption{An example of control results when applying each method}
\label{distance}
\end{figure}

Figure \ref{distance} shows an example of the control results when each method was applied to two tasks for the same initial position.
The results show that \textbf{ViT-B/32} could not accomplish either task, whereas \textbf{ViT-L/14} could only accomplish the box arrangement.
Furthermore, \textbf{ViT-B/32 (finetune)} completed both tasks.

Overall, we confirmed that the proposed method can control an object for different robots without learning control policies. Moreover, we improve the performance of our method by fine-tuning the CLIP model through a real experiment.

\begin{table}[!tp]
\caption{Comparisons of success rate for the real robot experiment (The target position of the object is unknown except for $\dagger$)}
\small
\vspace{1mm}
\label{eval1}
\centering
\renewcommand{\arraystretch}{1.1}
\begin{tabular}{ccccc}
\hline
Task &  \textbf{Goal}$\dagger$ & \textbf{ViT-B/32} & \textbf{ViT-L/14}  & \textbf{ViT-B/32 (finetune)} \\
\hline \hline
chair-rearrangement & (1.00) & 0.43 & 0.23 & \textbf{0.83} \\
box-rearrangement & (1.00) & 0.27 & 0.50 & \textbf{0.80} \\
\hline     
\end{tabular}
\end{table}

\begin{table}[!tp]
\caption{Comparison of text accuracy rates for the real images}
\small
\vspace{1mm}
\label{eval2}
\centering
\renewcommand{\arraystretch}{1.1}
\begin{tabular}{ccccc}
\hline
Task & \textbf{ViT-B/32} & \textbf{ViT-L/14}  & \textbf{ViT-B/32 (finetune)} \\
\hline \hline
chair-rearrangement & 0.66  & 0.50 & \textbf{0.89} \\
box-rearrangement & 0.48 & 0.73 & \textbf{0.95} \\
\hline     
\end{tabular}
\end{table}

\section{Discussion}
This section discusses the limitations of this study and future work.

Through a multitask simulation, it was confirmed that the proposed method could be applied to six tasks using the same CLIP model. 
However, the original CLIP could not achieve a high performance, and our method required fine-tuning for each task to improve performance. Therefore, a generalized CLIP model that can extract information about robot movement should be constructed.

Several assumptions were made in the simulations and experiments. In the simulation, we assumed that the position of the object handle was known because the current system could not make the end effector approach the handle. Therefore, an object handle should be detected using object detection methods such as YOLOv8 \cite{reis2023realtime} or Detic \cite{zhou2022detecting} and the position of the object handle should be estimated. Moreover, the effectiveness of the proposed method should be confirmed using the estimated object position. In the experiment, we assumed that the robot was rigidly attached to the object because the current method cannot determine which part of the object the robot should push. To address this issue, we combine our method with AffordanceNet \cite{deng20213d} to determine the pushing point of the object.

A possible future direction is to apply our method to robots equipped with onboard cameras. To this end, we should make our method robust to changes in camera position and angle by collecting several combinations of camera positions and angles and confirming the effectiveness of our approach. It would be interesting to apply our method to various types of robots other than two-wheeled robots and robotic arms. Furthermore, we may combine our method with VLMs other than the CLIP to examine which VLM model is the best suited for our method.

\section{Conclusion}
In this study, we proposed a CLIP-feature-based randomized control system without learning control policies. Our framework combines the vision-language CLIP model with a randomized control. In our method, the similarity between the camera images and text instructions was computed using CLIP. Moreover, the robot was controlled by a randomized control system that alternately repeated stochastic and deterministic movements. This renders our method applicable to multiple tasks and robots without learning the control policy. Through a multitask simulation and real robot experiment, we confirmed that our method using the original CLIP could achieve a success rate to some extent without learning the control policy for multiple tasks, and its performance was improved by fine-tuning the CLIP model.
In future work, we plan to improve the CLIP model to extract knowledge regarding robot movements and confirm the effectiveness of the proposed method for numerous robots and tasks.


\bibliographystyle{unsrt}  
\bibliography{references}  

\end{document}